\documentclass[10pt,twocolumn,letterpaper]{article}

\usepackage{cvpr}
\usepackage{times}
\usepackage{epsfig}
\usepackage{graphicx}
\usepackage{amsmath}
\usepackage{amssymb}

\usepackage{mathtools}
\usepackage{tabularx}
\newcolumntype{Y}{>{\centering\arraybackslash}X}
\usepackage{multirow}
\usepackage{enumitem}
\usepackage{subcaption}
\usepackage{printlen}
\usepackage{array}

\usepackage{textcomp}
\usepackage{xfrac}
\newcommand{\stimes}{{\times}}
\newcommand{\printsizefour}{\sfrac{D}{4}{\times}\sfrac{H}{4}{\times}\sfrac{W}{4}{\times}}
\newcommand{\printsizeeight}{\sfrac{D}{8}{\times}\sfrac{H}{8}{\times}\sfrac{W}{8}{\times}}
\newcommand{\printsizesixteen}{\sfrac{D}{16}{\times}\sfrac{H}{16}{\times}\sfrac{W}{16}{\times}}

\usepackage[pagebackref=true,breaklinks=true,letterpaper=true,colorlinks,bookmarks=false]{hyperref}

\cvprfinalcopy % *** Uncomment this line for the final submission

\def\cvprPaperID{1410} % *** Enter the CVPR Paper ID here

\ifcvprfinal\fi
\begin{document}

%%%%%%%%%%%%%%%%%%%%%%%%%%%%%%%%%%%%%%%%%%%%%%%%%%%%%%%%%%%%%
%%%%%%%%%%%%%%%%%%%%% TITLE & AUTHOR %%%%%%%%%%%%%%%%%%%%%%%%
%%%%%%%%%%%%%%%%%%%%%%%%%%%%%%%%%%%%%%%%%%%%%%%%%%%%%%%%%%%%%
\title{Group-wise Correlation Stereo Network}

\author{Xiaoyang Guo\\
Institution1\\
Institution1 address\\
{\tt\small firstauthor@i1.org}
\and
Second Author\\
Institution2\\
First line of institution2 address\\
{\tt\small secondauthor@i2.org}
}

\author{Xiaoyang Guo$^1$~~~~~
Kai Yang$^2$~~~~~
Wukui Yang$^2$~~~~~~
Xiaogang Wang$^1$~~~~~~
Hongsheng Li$^1$\\
$^1$ The Chinese University of Hong Kong~~~~~~~~~~~~~~~~~~~~~~~~~$^2$ SenseTime Research~~~~~~~~~ \\
{\tt\small \{xyguo, xgwang, hsli\}@ee.cuhk.edu.hk}~~~~~~~~~~~{\tt\small \{yangkai, yangwukui\}@sensetime.com}
}

%\author{Mingyang~Liang$^{1,2}$\footnotemark[1], Xiaoyang~Guo$^{3}$\thanks{These authors contributed equally to this work.}, Hongsheng~Li{$^3$}, Xiaogang~Wang{$^{3}$}, You~Song$^{1}$ \thanks{Corresponding author.} \\
%{$^1$Beihang University, Beijing, China}\\
%{$^2$SenseTime Research}\\
%{$^3$The Chinese University of Hong Kong, Hong Kong, China}\\
%{\{liangmingyang,songyou\}@buaa.edu.cn, \{xyguo, hsli, xgwang\}@ee.cuhk.edu.hk}
%}

\maketitle
%\thispagestyle{empty}

%%%%%%%%%%%%%%%%%%%%%%%%%%%%%%%%%%%%%%%%%%%%%%%%%%%%%%%%%%%%%
%%%%%%%%%%%%%%%%%%%%%%%%% ABSTRACT %%%%%%%%%%%%%%%%%%%%%%%%%%
%%%%%%%%%%%%%%%%%%%%%%%%%%%%%%%%%%%%%%%%%%%%%%%%%%%%%%%%%%%%%
\begin{abstract}
Stereo matching estimates the disparity between a rectified image pair, which is of great importance to depth sensing, autonomous driving, and other related tasks. Previous works built cost volumes with cross-correlation or concatenation of left and right features across all disparity levels, and then a 2D or 3D convolutional neural network is utilized to regress the disparity maps. In this paper, we propose to construct the cost volume by group-wise correlation. The left features and the right features are divided into groups along the channel dimension, and correlation maps are computed among each group to obtain multiple matching cost proposals, which are then packed into a cost volume. Group-wise correlation provides efficient representations for measuring feature similarities and will not lose too much information like full correlation. It also preserves better performance when reducing parameters compared with previous methods. The 3D stacked hourglass network proposed in previous works is improved to boost the performance and decrease the inference computational cost. Experiment results show that our method outperforms previous methods on Scene Flow, KITTI 2012, and KITTI 2015 datasets. The code is available at \url{https://github.com/xy-guo/GwcNet}
\end{abstract}

%%%%%%%%%%%%%%%%%%%%%%%%%%%%%%%%%%%%%%%%%%%%%%%%%%%%%%%%%%%%%
%%%%%%%%%%%%%%%%%%%%%%%%% Introduction
%%%%%%%%%%%%%%%%%%%%%%%%%%%%%%%%%%%%%%%%%%%%%%%%%%%%%%%%%%%%%
\section{Introduction}
% intro, The application of stereo matching, such as autonomous driving
Accurate depth sensing is the core of many computer vision applications, such as autonomous driving, robot navigation, and shallow depth-of-field image synthesis. Stereo matching belongs to passive depth sensing techniques, which estimates depth by matching pixels from rectified image pairs captured by two cameras. The disparity $d$ of a pixel can be converted into depth by $Fl/d$, where $F$ denotes the focal length of the camera lens, and $l$ is the distance between two camera centers. Therefore, the depth precision improves with the precision of disparity prediction.

% Existing stereo matching algorithms? The method details?
Traditional stereo pipelines usually consist of all or portion of the following four steps, matching cost computation, cost aggregation, disparity optimization, and post-processing~\cite{scharstein2002taxonomy}. Matching cost computation provides initial similarity measures for left image patches and possible corresponding right image patches, which is a crucial step of stereo matching. Some common matching costs include absolute difference (SAD), sum of squared difference (SSD), and normalized cross-correlation (NCC). The cost aggregation and optimization steps incorporate contextual matching costs and regularization terms to obtain more robust disparity predictions.

% recent algorithms
Learning-based methods explore different feature representations and aggregation algorithms for matching costs. DispNetC~\cite{mayer2016large} computes a correlation volume from the left and right image features and utilizes a CNN to directly regress disparity maps. GC-Net~\cite{kendall2017end} and PSMNet~\cite{chang2018pyramid} construct concatenation-based feature volume and incorporate a 3D CNN to aggregate contextual features. There are also works~\cite{batsos2018cbmv,schonberger2018learning} trying to aggregate evidence from multiple hand-crafted matching cost proposals. However, the above methods have several drawbacks. The full correlation~\cite{mayer2016large} provides an efficient way for measuring feature similarities, but it loses much information because it produces only a single-channel correlation map for each disparity level. The concatenation volume~\cite{kendall2017end,chang2018pyramid} requires more parameters in the following aggregation network to learn the similarity measurement function from scratch. \cite{batsos2018cbmv,schonberger2018learning} stills utilizes traditional matching costs and cannot be optimized end-to-end.

%the idea of the traditional stereo matching algorithms into 3D convolution networks and achieved great success. Instead of operating on raw images, the unary features $\mathbf{f}_l, \mathbf{f}_r$ of the left and the right images are first extracted with an shared-parameter feature encoder. Then, a 3D matching volume is constructed by concatenating the left image feature and the corresponding right image feature for all possible disparities.
%A 3D convolutional network is employed to aggregate contextual information for the feature volume to predict aggregated matching costs and regress disparity maps. However, the 3D network has to learn matching metrics from concatenated features. To save memory usage and make the computational cost of the 3D CNN tractable, the unary features are usually compressed into only a few channels, for example from 256 to 32 channels, which prevents the 3D network from fully utilizing the representation capability of the features.

In this paper, we propose a simple yet efficient operation called group-wise correlation to tackle the above drawbacks. Multi-level unary features are extracted and concatenated to form high-dimensional feature representations $\mathbf{f}_l,\mathbf{f}_r$ for a left-right image pair. Then, the features are split into multiple groups along the channel dimension, and the $i$th left feature group is cross-correlated with the corresponding $i$th right feature group over all disparity levels to obtain group-wise correlation maps. At last, all the correlation maps are packed to form the final 4D cost volume. The unary features can be treated as groups of structured vectors~\cite{wu2018group}, so the correlation maps for a certain group can be seen as a matching cost proposal. In this way, we can leverage the power of traditional cross-correlation matching cost and provide better similarity measures for the following 3D aggregation network compared with \cite{kendall2017end,chang2018pyramid}. The multiple matching cost proposals also avoid the information loss like full correlation~\cite{mayer2016large}.

The 3D stacked hourglass aggregation network proposed in PSMNet~\cite{chang2018pyramid} is modified to further improve the performance and decrease the inference computational cost. $1\stimes1\stimes1$ 3D convolutions are employed in the shortcut connections within each hourglass module without increasing too much computational cost.

% summary
Our main contributions can be summarized as follows.
1) We propose group-wise correlation to construct cost volumes to provide better similarity measures.
2) The stacked 3D hourglass refinement network is modified to improve the performance without increasing the inference time.
3) Our method achieves better performance than previous methods on Scene Flow, KITTI 2012, and KITTI 2015 datasets.
4) Experiment results show that when limiting the computational cost of the 3D aggregation network, the performance reduction of our proposed network is much smaller than previous PSMNet, which makes group-wise correlation a valuable way to be implemented in real-time stereo networks.

%%%%%%%%%%%%%%%%%%%%%%%%%%%%%%%%%%%%%%%%%%%%%%%%%%%%%%%%%%%%%
%%%%%%%%%%%%%%%%%%%%%%%%% Related Work
%%%%%%%%%%%%%%%%%%%%%%%%%%%%%%%%%%%%%%%%%%%%%%%%%%%%%%%%%%%%%
\section{Related Work}

\subsection{Traditional methods}
Generally, traditional stereo matching consists of all or portion of the following four steps: matching cost computation, cost aggregation, disparity optimization, and some post-processing steps~\cite{scharstein2002taxonomy}. In the first step, the matching costs of all pixels are computed for all possible disparities. Common matching costs include sum of absolute difference (SAD), sum of squared difference (SSD), normalized cross-correlation (NCC), and so on. Local methods~\cite{zhang2009cross,yang2012non,mei2013segment} explore different strategies to aggregate matching costs with neighbor pixels and usually utilize the winner-take-all (WTA) strategy to choose the disparity with minimum matching cost. In contrast, global methods minimize a target function to solve the optimal disparity map, which usually takes both matching costs and smoothness priors into consideration, such as belief propagation~\cite{sun2003stereo,klaus2006segment} and graph cut~\cite{kolmogorov2001computing}. Semi-global matching (SGM)~\cite{hirschmuller2005accurate} approximates the global optimization with dynamic programming. Local and global methods can be combined to obtain better performance and robustness.

\subsection{Learning based methods}
Besides hand-crafted methods, researchers also proposed many learned matching costs~\cite{zbontar2015computing,luo2016efficient,shaked2017improved} and cost aggregation algorithms~\cite{batsos2018cbmv,schonberger2018learning}. Zbontar and Lecun~\cite{zbontar2015computing} first proposed to compute matching costs using neural networks. The predicted matching costs are then processed with traditional cross-based cost aggregation and semi-global matching to predict the disparity map. The matching cost computation was accelerated in \cite{luo2016efficient} by correlating unary features. Batsos \etal proposed CBMV~\cite{batsos2018cbmv} to combine evidence from multiple basic matching costs. Schonberger \etal~\cite{schonberger2018learning} proposed to classify scanline matching cost candidates with a random forest classifier. Seki \etal proposed SGM-Nets~\cite{seki2017sgm} to provide learned penalties for SGM. Knobelreiter \etal~\cite{knobelreiter2017end} proposed to combine CNN-predicted correlation matching costs and CRF to integrate long-range interactions.

Following DispNetC (Mayer \etal~\cite{mayer2016large}), there are a lot of works directly regressing disparity maps from correlation cost volumes~\cite{pang2017cascade,liang2018learning,song2018edgestereo,yang2018segstereo}. Given the left and the right feature maps $\mathbf{f}_l$ and $\mathbf{f}_r$, the correlation cost volume is computed for each disparity level $d$,
\begin{equation}
    \textbf{C}_{corr}(d,x,y)=\frac{1}{N_c} \langle \mathbf{f}_l(x,y), \mathbf{f}_{r}(x-d,y) \rangle,
\end{equation}
where $\langle\cdot,\cdot\rangle$ is the inner product of two feature vectors and $N_c$ denotes the number of channels. CRL~\cite{pang2017cascade} and iResNet~\cite{liang2018learning} followed the idea of DispNetC with stack refinement sub-networks to further improve the performance. There are also works integrating additional information such as edge features~\cite{song2018edgestereo} and semantic features~\cite{yang2018segstereo}.

Recent works employed concatenation-based feature volume and 3D aggregation networks for better context aggregation~\cite{kendall2017end,chang2018pyramid,yu2018deep}. Kendall \etal proposed GC-Net~\cite{kendall2017end} and was the first to use 3D convolution networks to aggregate context for cost volumes. Instead of directly giving a cost volume, the left and the right feature $\mathbf{f}_l$, $\mathbf{f}_r$ are concatenated to form a 4D feature volume,
\begin{equation}
    \textbf{C}_{concat}(d,x,y,\cdot)=\text{Concat}\left\{\mathbf{f}_l(x,y), \mathbf{f}_{r}(x-d,y)\right\}.
\end{equation}
Context features are aggregated from neighbour pixels and disparities with 3D convolution networks to predict a disparity probability volume. Following GC-Net, Chang \etal~\cite{chang2018pyramid} proposed the pyramid stereo matching network (PSMNet) with a spatial pyramid pooling module and stacked 3D hourglass networks for cost volume refinement. Yu \etal~\cite{yu2018deep} proposed to generate and select multiple cost aggregation
proposals. Zhong \etal~\cite{zhong2018open} proposed a self-adaptive recurrent stereo model to tackle open-world video data.

LRCR~\cite{jie2018left} utilized left-right consistency check and recurrent model to aggregate cost volumes predicted from~\cite{shaked2017improved} and refined unreliable disparity predictions. There are also other works focusing on real-time stereo matching~\cite{khamis2018stereonet} and application friendly stereo~\cite{tulyakov2018practical}. 

%%%%%%%%%%%%%%%%%%%%%%%%%%%%%%%%%%%%%%%%%%%%%%%%%%%%%%%%%%%%%
%%%%%%%%%%%%%%%%%%%%%%%%%% Method
%%%%%%%%%%%%%%%%%%%%%%%%%%%%%%%%%%%%%%%%%%%%%%%%%%%%%%%%%%%%%

\begin{figure*}[t!]
\centering
\includegraphics[width=\linewidth]{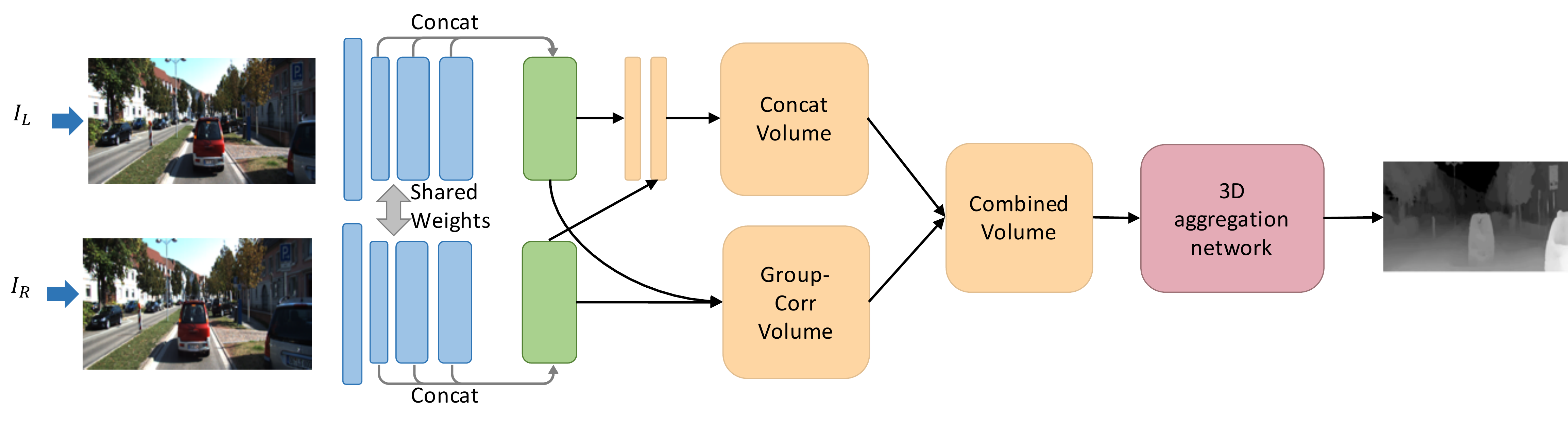}
\caption{The pipeline of the proposed group-wise correlation network. The whole network consists of four parts, unary feature extraction, cost volume construction, 3D convolution aggregation, and disparity prediction. The cost volume is divided into two parts, concatenation volume (\textit{Cat}) and group-wise correlation volume (\textit{Gwc}). Concatenation volume is built by concatenating the compressed left and right features. Group-wise correlation volume is described in Section~\ref{sec:groupwise-correlation}.}
\label{fig:pipeline}
\end{figure*}

\begin{figure*}[t!]
\centering
\includegraphics[width=0.95\linewidth]{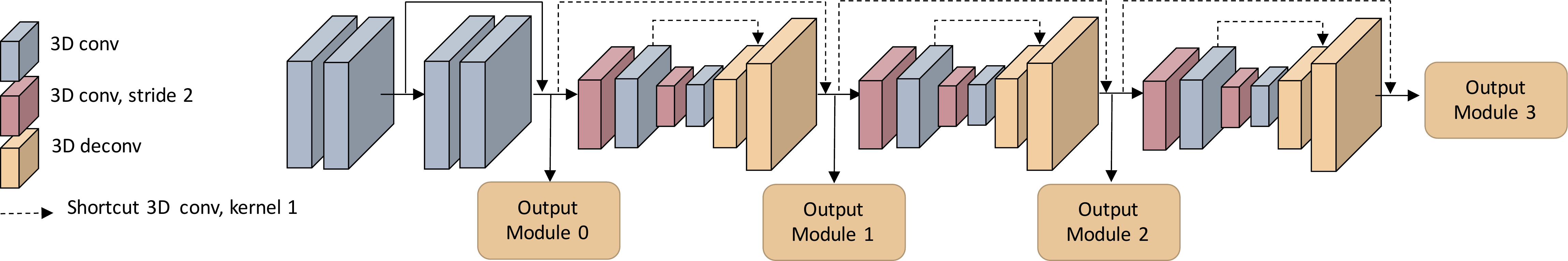}
\caption{The structure of our proposed 3D aggregation network. The network consists of a pre-hourglass module (four convolutions at the beginning) and three stacked 3D hourglass networks. Compared with PSMNet~\cite{chang2018pyramid}, we remove the shortcut connections between different hourglass modules and output modules, thus output modules 0,1,2 can be removed during inference to save time. $1\stimes1\stimes1$ 3D convolutions are added to the shortcut connections within hourglass modules.}
\label{fig:stackhourglass}
\end{figure*}

\section{Group-wise Correlation Network}
We propose group-wise correlation stereo network (GwcNet), which extends PSMNet~\cite{chang2018pyramid} with group-wise correlation cost volume and improved 3D stacked hourglass networks. In PSMNet, the matching costs for the concatenated features have to be learned from scratch by the 3D aggregation network, which usually requires more parameters and computational cost. In contrast, full correlation~\cite{mayer2016large} provides an efficient way for measuring feature similarities via dot product, but it loses much information. Our proposed group-wise correlation overcomes both drawbacks and provides good features for similarity measures.

\subsection{Network architecture}
The structure of the proposed group-wise correlation network is shown in Figure~\ref{fig:pipeline}. The network consists of four parts, unary feature extraction, cost volume construction, 3D aggregation, and disparity prediction. The detailed structures of the cost volume, stacked hourglass, and output modules are listed in Table~\ref{table:structure-details}.

For feature extraction, we adopt the ResNet-like network used in PSMNet~\cite{chang2018pyramid} with the half dilation settings and without its spatial pyramid pooling module.
The last feature maps of conv2, conv3, and conv4 are concatenated to form 320-channel unary feature maps.

The cost volume is composed of two parts, a concatenation volume and a group-wise correlation volume. The concatenation volume is the same as PSMNet~\cite{chang2018pyramid} but with fewer channels, before which the unary features are compressed into 12 channels with two convolutions. The proposed group-wise correlation volume will be described in details in Section \ref{sec:groupwise-correlation}. The two volumes are then concatenated as the input to the 3D aggregation network.

The 3D aggregation network is used to aggregate features from neighboring disparities and pixels to predict refined cost volumes. It consists of a pre-hourglass module and three stacked 3D hourglass networks to regularize the feature volumes. As shown in Figure~\ref{fig:stackhourglass}, the pre-hourglass module consists of four 3D convolutions with batch normalization and ReLU. Three stacked 3D hourglass networks are followed to refine low-texture ambiguities and occlusion parts by encoder-decoder structures. Compared with 3D aggregation network of \cite{chang2018pyramid}, we have several important modifications to improve the performance and increase the inference speed, and details are described in Section~\ref{sec:modify-hourglass}.

The pre-hourglass module and three stacked 3D hourglass networks are connected to output modules. Each output module predicts a disparity map. The structure of the output module and the loss function are described in Section~\ref{sec:loss}.

\subsection{Group-wise correlation volume}
\label{sec:groupwise-correlation}
The left unary features and the right unary features are denoted by $\mathbf{f}_l$ and $\mathbf{f}_r$ with $N_c$ channels and in $1{/}4$ the size of original images. In previous works~\cite{mayer2016large,kendall2017end,chang2018pyramid}, the left and right features are correlated or concatenated at different disparity levels to form the cost volume. However, both correlation volume and concatenation volume have drawbacks. The full correlation provides an efficient way for measuring feature similarities, but it loses much information because it produces only a single-channel correlation map for each disparity level. The concatenation volume contains no information about the feature similarities, so more parameters are required in the following aggregation network to learn the similarity measurement function from scratch. To solve the above issues, we propose group-wise correlation by combining advantages of the concatenation volume and the correlation volume.

The basic idea behind group-wise correlation is splitting the features into groups and computing correlation maps group by group. We denote the channels of unary features as $N_c$. All the channels are evenly divided into $N_g$ groups along the channel dimension, and each feature group therefore has $N_c/N_g$ channels. The $g$th feature group $\mathbf{f}_l^g,\mathbf{f}_r^g$ consists of the $g\frac{N_c}{N_g}, g\frac{N_c}{N_g}+1, \dots, g\frac{N_c}{N_g}+(\frac{N_c}{N_g}-1)$th channels of the original feature $\mathbf{f}_l,\mathbf{f}_r$. The group-wise correlation is then computed as
\begin{equation}
  \textbf{C}_{gwc}(d,x,y,g)=\frac{1}{N_c/N_g} \langle \mathbf{f}_l^g(x,y), \mathbf{f}_{r}^g(x-d,y) \rangle,
\end{equation}
where $\langle\cdot,\cdot\rangle$ is the inner product. Note that the correlation is computed for all feature groups $g$ and at all disparity levels $d$. Then, all the correlation maps are packed into a matching cost volume of the shape $[D_{max}/4,H/4,W/4,N_g]$, where $D_{max}$ denotes the maximum disparity and $D_{max}/4$ corresponds to the maximum disparity for the feature. When $N_g{=}1$, the group-wise correlation becomes full correlation. 

Group-wise correlation volume $\textbf{C}_{gwc}$ can be treated as $N_g$ cost volume proposals, and each proposal is computed from the corresponding feature group. The following 3D aggregation network aggregates multiple candidates to regress disparity maps. The group-wise correlation successfully leverages the power of traditional correlation matching costs and provides rich similarity-measure features for the 3D aggregation network, which alleviates the parameter demand. We will show in Section~\ref{sec:exp-reduce-layer-channel} that we explore to reduce the channels of the 3D aggregation network, and the performance reduction of our proposed network is much smaller than \cite{chang2018pyramid}. Our proposed group-wise correlation volume requires less 3D aggregation parameters to achieve favorable results.

To further improve the performance, the group correlation cost volume can be combined with the concatenation volume. Experiment results show that the group-wise correlation volume and the concatenation volume are complementary to each other.

\begin{table}[t]
\begin{center}
\footnotesize
\begin{tabularx}{\linewidth}{@{\hskip 0.2em}l@{\hskip 0.3em}|@{\hskip 0.3em}X@{\hskip 0.3em}|@{\hskip 0.3em}l@{\hskip 0.2em}}
\hline
Name & Layer properties & Output size \\
\hline
\multicolumn{3}{c}{Cost Volume} \\
\hline
unary\_l/r & N/A, S2 & $\sfrac{H}{4}\stimes \sfrac{W}{4}\stimes 320$ \\
\hline
volume\_g & group-wise cost volume & $\printsizefour 40$ \\
\hline
volume\_c & concatenation cost volume & $\printsizefour 24$ \\
\hline
volume & volume\_g,volume\_c: Concat & $\printsizefour 64$ \\
\hline
\multicolumn{3}{c}{Pre-hourglass} \\
\hline
conv1 & [$32\stimes32$, $3\stimes3\stimes3$, S1] $\stimes 2$ & $\printsizefour 32$ \\
\hline
conv2 & [$32\stimes32$, $3\stimes3\stimes3$, S1] $\stimes 2$ & $\printsizefour 32$ \\
\hline
output & conv1,conv2: Add & $\printsizefour 32$ \\
\hline
\multicolumn{3}{c}{Hourglass Module 1, 2, 3} \\
\hline
input & N/A & $\printsizefour 32$ \\
\hline
conv1a & $32\stimes64$, $3\stimes3\stimes3$, S2 & $\printsizeeight 64$ \\
conv1b & $64\stimes64$, $3\stimes3\stimes3$, S1 & $\printsizeeight 64$ \\
\hline
conv2a & $64\stimes128$, $3\stimes3\stimes3$, S2 & $\printsizesixteen 128$ \\
conv2b & $128\stimes128$, $3\stimes3\stimes3$, S1 & $\printsizesixteen 128$ \\
\hline
deconv1* & $128\stimes64$, $3\stimes3\stimes3$, S2, deconv & $\printsizeeight 64$ \\
shortcut1* & conv1b: $64\stimes64$, $1\stimes1\stimes1$, S1 & $\printsizeeight 64$ \\
plus1 & deconv1,shortcut1: Add\&ReLU & $\printsizeeight 64$ \\
\hline
deconv0* & $64\stimes32$, $3\stimes3\stimes3$, S2, deconv & $\printsizefour 32$ \\
shortcut0* & input: $32\stimes32$, $1\stimes1\stimes1$, S1 & $\printsizefour 32$ \\
output & deconv0,shortcut0: Add\&ReLU & $\printsizefour 32$ \\
\hline
\multicolumn{3}{c}{Output Module 0, 1, 2, 3} \\
\hline
input & N/A & $\printsizefour 32$ \\
\hline
conv1 & $32\stimes32$, $3\stimes3\stimes3$, S1 & $\printsizefour 32$ \\
\hline
conv2** & $32\stimes1$, $3\stimes3\stimes3$, S1 & $\printsizefour 1$ \\
\hline
score & Upsample & $ D \stimes H \stimes W \stimes 1$ \\
\hline
prob & Softmax (at disparity dimension) & $D \stimes H \stimes W \stimes 1$ \\
\hline
disparity & Soft Argmin (Equ. \ref{equ:soft-argmin}) & $H \stimes W \stimes 1$ \\
\hline
\end{tabularx}
\end{center}
\caption{Structure details of the modules. $H,W$ represents the height and the width of the input image. S1/2 denotes the convolution stride. If not specified, each 3D convolution is with a batch normalization and ReLU. * denotes the ReLU is not included. ** denotes convolution only.}
\label{table:structure-details}
\end{table}

\subsection{Improved 3D aggregation module}
\label{sec:modify-hourglass}
In PSMNet~\cite{chang2018pyramid}, a stacked hourglass architecture was proposed to learn better context features. Based on the network, we apply several important modifications to make it suitable for our proposed group-wise correlation and improve the inference speed. The structure of the proposed 3D aggregation is shown in Figure~\ref{fig:stackhourglass} and Table~\ref{table:structure-details}.

% TODO: introduction to previous network structure?

First, we add one more auxiliary output module (output module 0, see Figure~\ref{fig:stackhourglass}) for the features of the pre-hourglass module. The extra auxiliary loss makes the network learn better features at lower layers, which also benefits the final prediction.

Second, the residual connections between different output modules are removed, thus auxiliary output modules (output module 0, 1, 2) can be removed during inference to save computational cost.

Third, $1\stimes1\stimes1$ 3D convolutions are added to the shortcut connections within each hourglass module (see dashed lines in Figure~\ref{fig:stackhourglass}) to improve the performance without increasing much computational cost. Since the $1\stimes1\stimes1$ 3D convolution only has 1/27 multiplication operations compared with $3\stimes3\stimes3$ convolutions, it runs very fast and the time can be neglected.

\begin{table*}[t!]
\begin{center}
\setlength{\tabcolsep}{2pt}
\small
\begin{tabularx}{1.0\linewidth}{@{}l|YYY|Y|Y|Y|Y|Y|Y|Y@{}}
\hline
Model & Concat Volume & Group Corr Volume & Stack Hourglass & Groups $\stimes$ Channels & Init Volume Channel & \textgreater 1px (\%) & \textgreater 2px (\%) & \textgreater 3px (\%)  & EPE (px) & Time (ms) \\
\hline
Cat64-Base & \checkmark & & & - & 64 & 12.78 & 8.05 & 6.33 & 1.308 & 117.1 \\
% \hline
Gwc1-Base & & \checkmark & & 1$\stimes$320 & 1 & 13.32 & 8.37 & 6.62 & 1.369 & 104.0 \\
Gwc10-Base & & \checkmark & & 10$\stimes$32 & 10 & 11.82 & 7.31 & 5.70 & 1.230 & 112.8\\
Gwc20-Base & & \checkmark & & 20$\stimes$16 & 20 & 11.84 & 7.29 & 5.67 & 1.216 & 116.3 \\
Gwc40-Base & & \checkmark & & 40$\stimes$8 & 40 & 11.68 & 7.18 & 5.58 & 1.212 & 122.2 \\
Gwc80-Base & & \checkmark & & 80$\stimes$4 & 80 & 11.69 & 7.17 & 5.57 & 1.214 & 133.3 \\
Gwc160-Base & & \checkmark & & 160$\stimes$2 & 160 & 11.58 & 7.08 & 5.49 & 1.188 & 157.3\\
% Gwc320-Base & Time209.9
% \hline
Gwc40-Cat24-Base & \checkmark & \checkmark & & 40$\stimes$8 & 40+24 & \textbf{11.26} & \textbf{6.87} & \textbf{5.31} & \textbf{1.127} & 135.1 \\
\hline
PSMNet~\cite{chang2018pyramid} & \checkmark & & \cite{chang2018pyramid} & - & 64 & 9.46 & 5.19 & 3.80 & 0.887 & 246.1 \\
Cat64-original-hg & \checkmark & & \cite{chang2018pyramid} & - & 64 & 9.47 & 5.13 & 3.74 & 0.876 & 241.0 \\
% Gwc40-Cat32 & \checkmark & \checkmark & \cite{chang2018pyramid} & 40$\stimes$8 & 40+32 & \\
Cat64 & \checkmark & & Ours & - & 64 & 8.41 & 4.63 & 3.41 & 0.808 & 198.3 \\
% Gwc40 & & \checkmark & \cite{chang2018pyramid} & 40$\stimes$8 & 40 & 9.30 & 5.09 & 3.74 & 0.879 & time\\
Gwc40 (GwcNet-g) & & \checkmark & Ours & 40$\stimes$8 & 40 & 8.18 & 4.57 & 3.39 & 0.792 & 200.3 \\
% \hline
% Gwc40-Cat32 & \checkmark & \checkmark & Ours & 40$\stimes$8 & 40+32 & 8.18 & 4.56 & 3.36 & 0.771 & 215.7 \\
Gwc40-Cat24 (GwcNet-gc) & \checkmark & \checkmark & Ours & 40$\stimes$8 & 40+24 & \textbf{8.03} & \textbf{4.47} & \textbf{3.30} & \textbf{0.765} & 210.7 \\
% Gwc40-Cat12 & \checkmark & \checkmark & Ours & 40$\stimes$8 & 40+12 & 8.06 & 4.47 & 3.31 & 0.765 & time \\
\hline

\end{tabularx}
\end{center}
\caption{Ablation study results of proposed networks on the Finalpass of Scene Flow datasets~\cite{mayer2016large}. \textit{Cat}, \textit{Gwc}, \textit{Gwc-Cat} represent only concatenation volume, only group-wise correlation volume, or the both. \textit{Base} denotes the network variants without stacked hourglass networks. The time is the inference time for 480$\stimes$640 inputs on a single Nvidia TITAN Xp GPU. The result of PSMNet~\cite{chang2018pyramid} is trained with published code with our batch size, evaluation settings for fair comparison.}
\label{table:scene-flow-results}
\end{table*}

\begin{table}[t]
\begin{center}
\setlength{\tabcolsep}{2pt}
\small
\begin{tabularx}{1.0\linewidth}{@{}l|Y|Y|Y|Y@{}}
\hline
Model & KITTI 12 EPE (px) & KITTI 12 D1-all(\%) & KITTI 15 EPE (px) & KITTI 15 D1-all (\%) \\
\hline
PSMNet~\cite{chang2018pyramid} & 0.713 & 2.53 & 0.639 & 1.50 \\
Cat64-original-hg & 0.740 & 2.72 & 0.652 & 1.76  \\
Cat64 & 0.691 & 2.41 & 0.615 & 1.55 \\
Gwc40 & 0.662 & 2.30 & \textbf{0.602} & \textbf{1.41}\\
Gwc40-Cat24 & \textbf{0.659} & \textbf{2.10} & 0.613 & 1.49 \\
\hline
\end{tabularx}
\end{center}
\caption{Ablation study results of our networks on KITTI 2012 validation and KITTI 2015 validation sets. }
\label{table:ablation-kitti}
\end{table}

\subsection{Output module and loss function}
\label{sec:loss}
For each output module, two 3D convolutions are employed to output a 1-channel 4D volume, and then the volume is upsampled and converted into a probability volume with softmax function along the disparity dimension. Detailed structures are shown in Table~\ref{table:structure-details}. For each pixel, we have a $D_{max}$-length vector which contains the probability $p$ for all disparity levels. Then, the disparity estimation $\widetilde{d}$ is given by the soft argmin function~\cite{kendall2017end},
\begin{equation}
	\widetilde{d}=\sum_{k=0}^{D_{max}-1}k\cdot p_k,
  \label{equ:soft-argmin}
\end{equation}
where $k$ and $p_k$ denote a possible disparity level and the corresponding probability. The predicted disparity maps from the four output modules are denoted as $\mathbf{\widetilde{d}}_0,\mathbf{\widetilde{d}}_1,\mathbf{\widetilde{d}}_2,\mathbf{\widetilde{d}}_3$. The final loss is given by,
\begin{equation}
	L=\sum_{i=0}^{i=3}\lambda_i\cdot\text{Smooth}_{L_1}(\mathbf{\widetilde{d}}_i-\mathbf{d}^*),
\end{equation}
where $\lambda_i$ denotes the coefficients for the $i$th disparity prediction and $\mathbf{d}^*$ represents the ground-truth disparity map. The smooth L1 loss is computed as follows,

\begin{equation}  
\text{Smooth}_{L_1}(x)=
\left\{  
    \begin{aligned}
        & 0.5x^2, & \text{if } |x|<1 \\
        & |x|-0.5, & \text{otherwise} \\
	\end{aligned}
\right.  
\end{equation}  

%%%%%%%%%%%%%%%%%%%%%%%%%%%%%%%%%%%%%%%%%%%%%%%%%%%%%%%%%%%%%
%%%%%%%%%%%%%%%%%%%%%%%%% Experiment
%%%%%%%%%%%%%%%%%%%%%%%%%%%%%%%%%%%%%%%%%%%%%%%%%%%%%%%%%%%%%

\section{Experiment}
In this section, we evaluate our proposed stereo models on Scene Flow datasets~\cite{mayer2016large} and the KITTI dataset~\cite{Geiger2012CVPR,menze2015object}. We present ablation studies to compare different models and different parameter settings. Datasets and implementation details are described in Section~\ref{sec:datasets} and Section~\ref{sec:implementation-details}. The effectiveness and the best settings of group-wise correlation are explored in Section~\ref{sec:exp-groupwise-corr}. The performance improvement of the new stacked hourglass module is discussed in Section~\ref{sec:exp-stack-hourglass}. We also explore the performance of group-wise correlation when the computational cost is limited in Section~\ref{sec:exp-reduce-layer-channel}.

\subsection{Datasets and evaluation metrics}
\label{sec:datasets}

\textbf{Scene Flow datasets} are a dataset collection of synthetic stereo datasets, consisting of Flyingthings3D, Driving, and Monkaa. The datasets provide 35,454 training and 4,370 testing images of size 960$\stimes$540 with accurate ground-truth disparity maps. We use the Finalpass of the Scene Flow datasets, since it contains more motion blur and defocus and is more like real-world images than the Cleanpass. \textbf{KITTI 2012} and \textbf{KITTI 2015} are driving scene datasets. KITTI 2012 provides 194 training and 195 testing images pairs, and KITTI 2015 provides 200 training and 200 testing image pairs. Both datasets provide sparse LIDAR ground-truth disparity for the training images.

For Scene Flow datasets, the evaluation metrics is usually the end-point error (EPE), which is the mean average disparity error in pixels. For KITTI 2012, percentages of erroneous pixels and average end-point errors for both non-occluded (Noc) and all (All) pixels are reported. For KITTI 2015, the percentage of disparity outliers \textit{D1} is evaluated for background, foreground, and all pixels. The outliers are defined as the pixels whose disparity errors are larger than $\text{max}(3\text{px}, 0.05d^*)$, where $d^*$ denotes the ground-truth disparity.

\subsection{Implementation details}
\label{sec:implementation-details}
Our network is implemented with PyTorch. We use Adam~\cite{kingma2014adam} optimizer, with $\beta_1=0.9$, $\beta_2=0.999$. The batch size is fixed to 16, and we train all the networks with 8 Nvidia TITAN Xp GPUs with 2 training samples on each GPU. The coefficients of four outputs are set as $\lambda_0=0.5$, $\lambda_1=0.5$, $\lambda_2=0.7$, $\lambda_3=1.0$.

For Scene Flow datasets, we train the stereo networks for 16 epochs in total. The initial learning rate is set to 0.001. It is down-scaled by 2 after epoch 10, 12, 14 and ends at 0.000125. To test on Scene Flow datasets, the full images of size 960$\stimes$540 are input to the network for disparity prediction. We set the maximum disparity value as $D_{max}=192$ following PSMNet~\cite{chang2018pyramid} for Scene Flow datasets. To evaluate our networks, we remove all the images with less than 10\% valid pixels ($0{\leq} d{<}D_{max}$) in the test set. For each valid image, the evaluation metrics are computed with only valid pixels.

For KITTI 2015 and KITTI 2012, we fine-tune the network pre-trained on Scene Flow datasets for another 300 epochs. The initial learning rate is 0.001 and is down-scaled by 10 after epoch 200. For testing on KITTI datasets, we first pad zeros on the top and the right side of the images to make the inputs in size 1248$\stimes$384.

% Ablation study 1
\subsection{The effectiveness of Group-wise correlation}
\label{sec:exp-groupwise-corr}
In this section, we explore the effectiveness and the best settings for the group-wise correlation. In order to prove the effectiveness of the proposed group-wise correlation volume, we conduct several experiments on the \textit{Base} model, which removes the stacked hourglass networks and only preserves the pre-hourglass module and the output module 0. \textit{Cat-Base}, \textit{Gwc-Base}, and \textit{Gwc-Cat-Base} are the base models with only concatenation volume, only group-wise correlation volume, or both volumes.

Experiment results in Table~\ref{table:scene-flow-results} show that the performance of the \textit{Gwc-Base} network increases as the group number increases. When the group number is larger than 40, the performance improvement becomes minor and the end-point error stays around 1.2px. Considering the memory usage and the computational cost, we choose 40 groups with each group having 8 channels as our network structure, which corresponds to the \textit{Gwc40-Base} model in Table~\ref{table:scene-flow-results}.

All the \textit{Gwc-Base} models except \textit{Gwc1-Base} outperform the \textit{Cat-Base} model which utilizes concatenation volume, which shows the effectiveness of the group-wise correlation. The \textit{Gwc40} model reduces the end-point error by 0.1px and the 3-pixel error rate by 0.75\%, and the time consumption is almost the same. The performance can be further improved by combining group-wise correlation volume with concatenation  volume (see \textit{Gwc40-Cat24-Base} model in Table~\ref{table:scene-flow-results}). The group-wise correlation could provide accurate matching features, and the concatenation volume provides complementary semantic information.

\begin{figure}[tbp]
\centering
\includegraphics[width=0.9\linewidth]{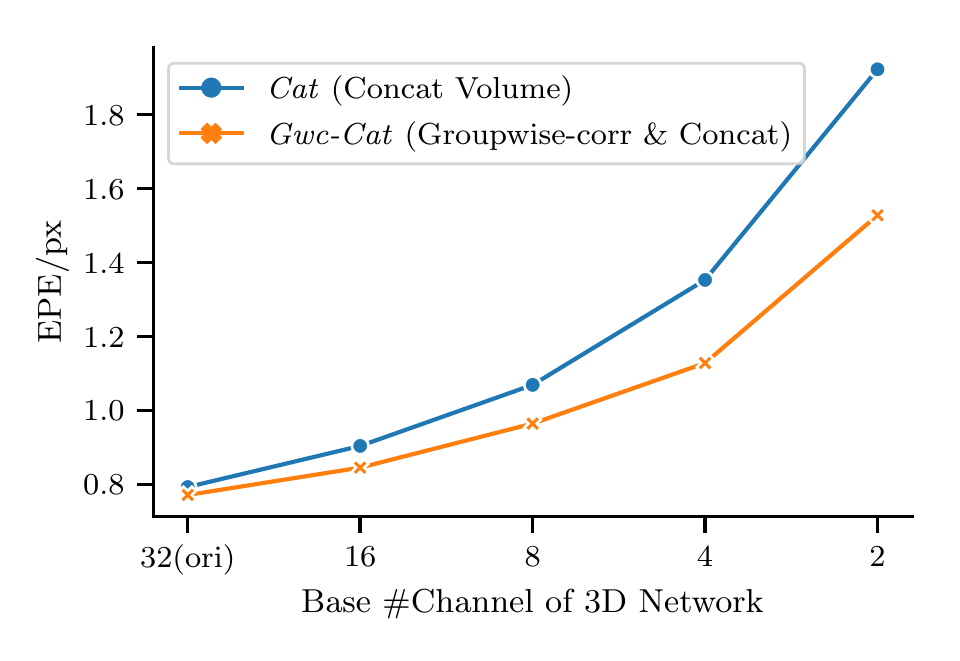}
\caption{Our model \textit{Gwc-Cat} achieves much better performance than \textit{Cat} when the number of channels decreases. The models with 32 base channels correspond to the \textit{Cat64} model (concatenation volume) and the \textit{Gwc40-Cat24} model (group-wise correlation and concatenation volume). The channels of the cost volume and all 3D convolutions decrease by the same factor as the base channel.}
 % When the ratio equals to $1/16$, the base channels are only 2, and group-correlation outperforms concat volume by 0.53px.
\label{fig:reduce-layer-channel}
\end{figure}

\begin{figure*}[t]
\begin{center}
\small
\setlength{\tabcolsep}{1pt}

  \begin{subfigure}[b]{\textwidth}
    \begin{tabularx}{\linewidth}{@{}YYY@{}}
      \includegraphics[width=\linewidth,height=0.45\linewidth]{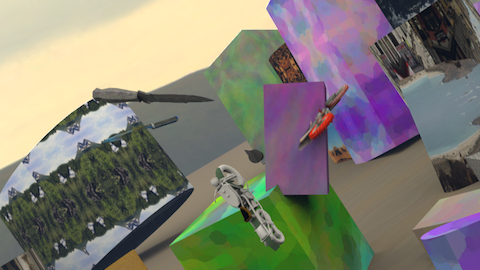} &
      \includegraphics[width=\linewidth,height=0.45\linewidth]{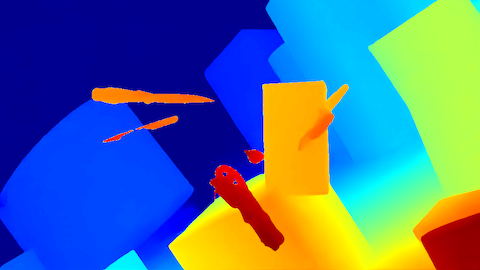} &
      \includegraphics[width=\linewidth,height=0.45\linewidth]{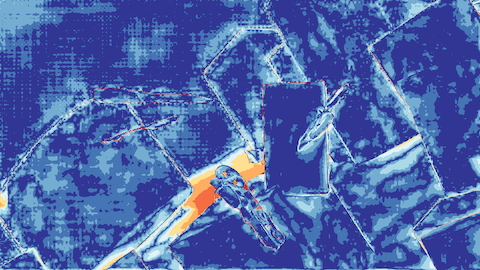} \\

      \includegraphics[width=\linewidth,height=0.45\linewidth]{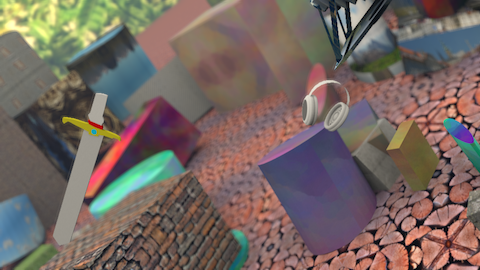} &
      \includegraphics[width=\linewidth,height=0.45\linewidth]{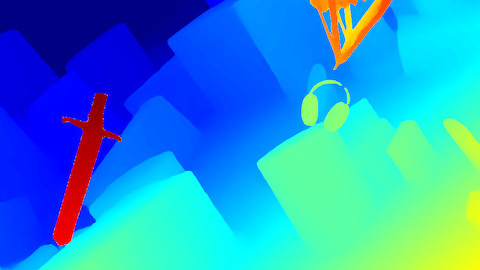} &
      \includegraphics[width=\linewidth,height=0.45\linewidth]{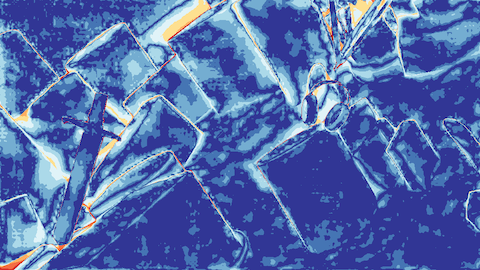} \\

      \includegraphics[width=\linewidth,height=0.45\linewidth]{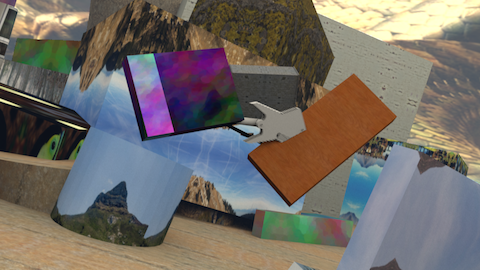} &
      \includegraphics[width=\linewidth,height=0.45\linewidth]{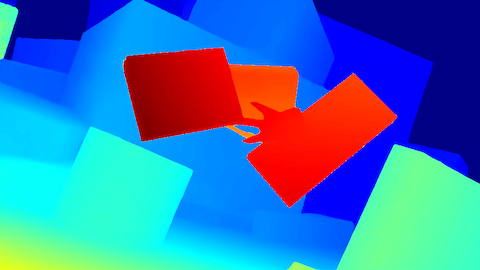} &
      \includegraphics[width=\linewidth,height=0.45\linewidth]{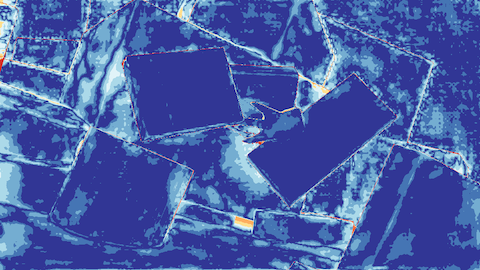} \\
    \end{tabularx}
    \caption{Visualization results on the Scene Flow datasets.}
    \label{fig:visualization-sceneflow}
  \end{subfigure}

  \begin{subfigure}[b]{\textwidth}
    \begin{tabularx}{\linewidth}{@{}YYY@{}}
      \includegraphics[width=\linewidth,height=0.26\linewidth]{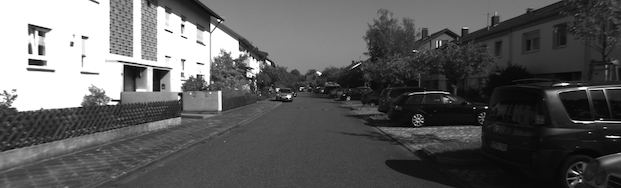} &
      \includegraphics[width=\linewidth,height=0.26\linewidth]{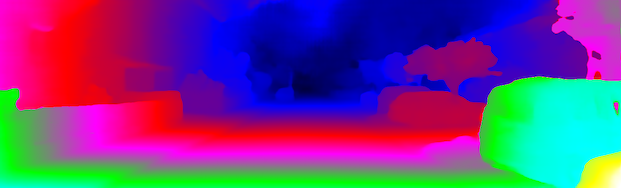} &
      \includegraphics[width=\linewidth,height=0.26\linewidth]{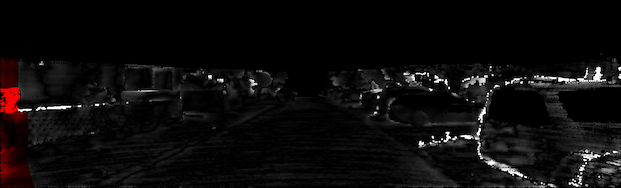} \\

      \includegraphics[width=\linewidth,height=0.26\linewidth]{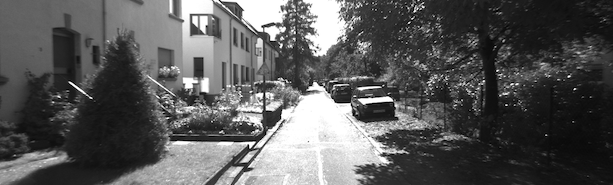} &
      \includegraphics[width=\linewidth,height=0.26\linewidth]{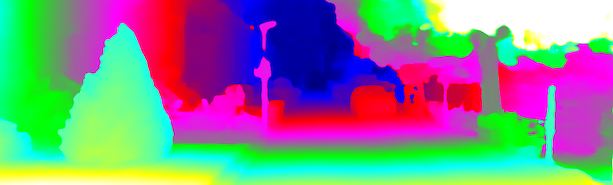} &
      \includegraphics[width=\linewidth,height=0.26\linewidth]{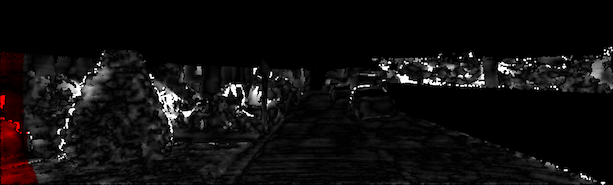} \\

      \includegraphics[width=\linewidth,height=0.26\linewidth]{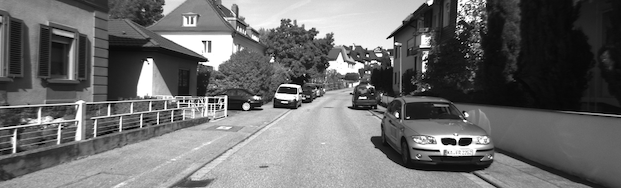} &
      \includegraphics[width=\linewidth,height=0.26\linewidth]{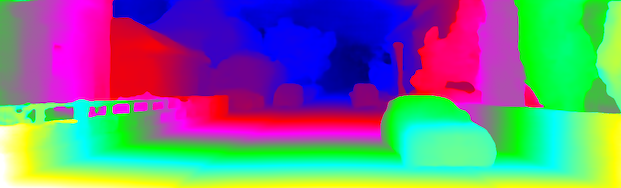} &
      \includegraphics[width=\linewidth,height=0.26\linewidth]{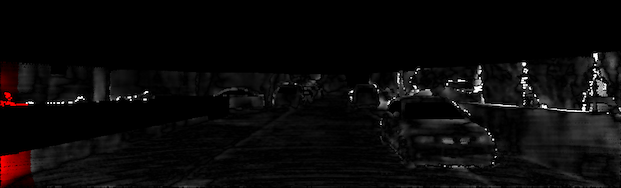} \\
    \end{tabularx}
    \caption{Visualization results on the KITTI 2012 dataset.}
    \label{fig:visualization-kitti2012}
  \end{subfigure}

  \begin{subfigure}[b]{\textwidth}
    \begin{tabularx}{\linewidth}{@{}YYY@{}}
      \includegraphics[width=\linewidth,height=0.3\linewidth]{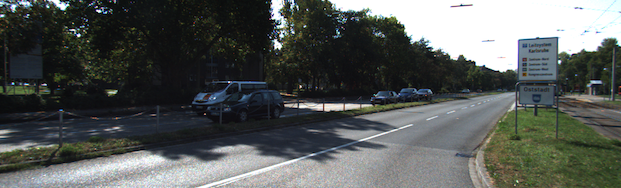} &
      \includegraphics[width=\linewidth,height=0.3\linewidth]{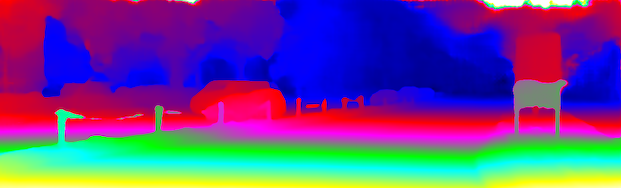} &
      \includegraphics[width=\linewidth,height=0.3\linewidth]{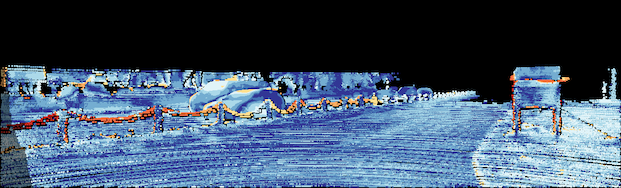} \\

      \includegraphics[width=\linewidth,height=0.3\linewidth]{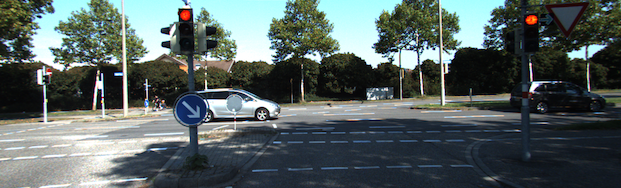} &
      \includegraphics[width=\linewidth,height=0.3\linewidth]{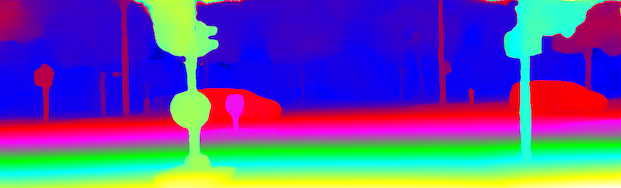} &
      \includegraphics[width=\linewidth,height=0.3\linewidth]{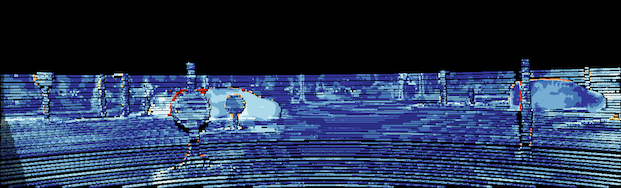} \\

      \includegraphics[width=\linewidth,height=0.3\linewidth]{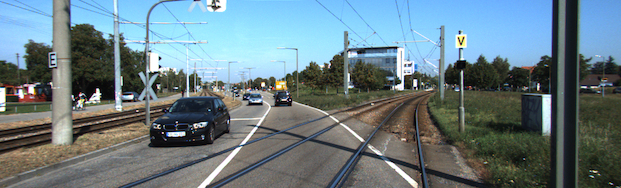} &
      \includegraphics[width=\linewidth,height=0.3\linewidth]{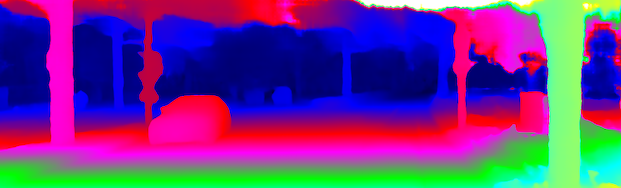} &
      \includegraphics[width=\linewidth,height=0.3\linewidth]{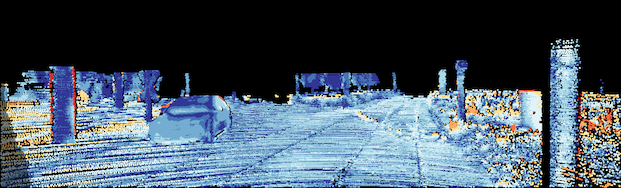} \\
    \end{tabularx}
    \caption{Visualization results on the KITTI 2015 dataset.}
    \label{fig:visualization-kitti2015}
  \end{subfigure}

\end{center}
\caption{Depth visualization results on the test sets of Scene Flow~\cite{mayer2016large}, KITTI 2012~\cite{Geiger2012CVPR} and KITTI 2015~\cite{menze2015object} datasets. From left to right, input left images, predicted disparity maps, and error maps.}
\end{figure*}

\begin{table*}[t]
\begin{center}
\begin{tabularx}{0.95\linewidth}{l|Y|Y|Y|Y|Y|Y|Y}
\hline
 & \multicolumn{3}{c|}{All (\%)} & \multicolumn{3}{c|}{Noc (\%)} & Time \\
 \cline{2-7}
 & D1-bg & D1-fg & D1-all & D1-bg & D1-fg & D1-all & (s) \\
\hline
DispNetC~\cite{mayer2016large} & 4.32 & 4.41 & 4.34 & 4.11 & 3.72 & 4.05 & 0.06 \\
GC-Net~\cite{kendall2017end} & 2.21 & 6.16 & 2.87 & 2.02 & 5.58 & 2.61 & 0.9 \\
CRL~\cite{pang2017cascade} & 2.48 & 3.59 & 2.67 & 2.32 & 3.12 & 2.45 & 0.47 \\
iResNet-i2e2~\cite{liang2018learning} & 2.14 & 3.45 & 2.36 & 1.94 & 3.20 & 2.15 & 0.22 \\
PSMNet~\cite{kendall2017end} & 1.86 & 4.62 & 2.32 & 1.71 & 4.31 & 2.14 & 0.41 \\
SegStereo~\cite{yang2018segstereo} & 1.88 & 4.07 & 2.25 & 1.76 & 3.70 & 2.08 & 0.6\\
\hline
GwcNet-g (Gwc40) & \textbf{1.74} & \textbf{3.93} & \textbf{2.11} & \textbf{1.61} & \textbf{3.49} & \textbf{1.92} & 0.32 \\
\hline
\end{tabularx}
\end{center}
\caption{KITTI 2015 test set results. The dataset contains 200 images for training and 200 images for testing. }
\label{table:result-kitti2015}
\end{table*}

\begin{table*}[t]
\begin{center}
\begin{tabularx}{0.95\linewidth}{l|Y|Y|Y|Y|Y|Y|Y|Y|Y}
\hline
 & \multicolumn{2}{c|}{\textgreater 2px (\%)} & \multicolumn{2}{c|}{\textgreater 3px (\%)} & \multicolumn{2}{c|}{\textgreater 5px (\%)}  & \multicolumn{2}{c|}{Mean Error (px)}  & Time \\
 \cline{2-9}
 & Noc & All & Noc & All & Noc & All & Noc & All & (s) \\
\hline
DispNetC~\cite{mayer2016large} & 7.38 & 8.11 & 4.11 & 4.65 & 2.05 & 2.39 & 0.9 & 1.0 & 0.06 \\
MC-CNN-acrt~\cite{zbontar2015computing} & 3.90 & 5.45 & 2.43 & 3.63 & 1.64 & 2.39 & 0.7 & 0.9 & 67 \\
GC-Net~\cite{kendall2017end} & 2.71 & 3.46 & 1.77 & 2.30 & 1.12 & 1.46 & 0.6 & 0.7 & 0.9 \\
iResNet-i2~\cite{liang2018learning} & 2.69 & 3.34 & 1.71 & 2.16 & 1.06 & 1.32 & 0.5 & 0.6 & 0.12 \\
SegStereo~\cite{yang2018segstereo} & 2.66 & 3.19 & 1.68 & 2.03 & 1.00 & 1.21 &  0.5 & 0.6 & 0.6 \\
PSMNet~\cite{kendall2017end} & 2.44 & 3.01 & 1.49 & 1.89 & 0.90 & 1.15 & 0.5 & 0.6 & 0.41\\
\hline
GwcNet-gc (Gwc40-Cat24) & \textbf{2.16} & \textbf{2.71} & \textbf{1.32} &	\textbf{1.70} & \textbf{0.80} &	\textbf{1.03} & \textbf{0.5} & \textbf{0.5} & 0.32  \\
\hline
\end{tabularx}
\end{center}
\caption{KITTI 2012 test set results. The dataset contains 194 images for training and 195 images for testing. }
\label{table:result-kitti2012}
\end{table*}

% TODO: BEGIN HERE
\subsection{Improved stacked hourglass}
\label{sec:exp-stack-hourglass}
In this paper, we applied several modifications to the stacked hourglass networks proposed in \cite{chang2018pyramid} to improve the performance of cost volume aggregation. From Table~\ref{table:scene-flow-results} and Table~\ref{table:ablation-kitti}, we can see that the model with the proposed hourglass networks (\textit{Cat64}) increases EPE by 7.8\% on Scene Flow datasets and 5.8\% on KITTI 2015 compared with the model \textit{Cat64-original-hg} (with the hourglass module in \cite{chang2018pyramid}). The inference time for $640\stimes480$ inputs on a single Nvidia TITAN Xp GPU also decreases by 42.7ms, because the auxiliary output modules can be removed during inference to save time.

\subsection{Limit the computational cost of 3D network}
\label{sec:exp-reduce-layer-channel}
We explore to limit the computational cost by decreasing channels in the 3D aggregation network to verify the effectiveness of the proposed group-wise group correlation. The results are shown in Figure~\ref{fig:reduce-layer-channel}. The base number of channels are modified from the original 32 to 2, and the channels of the cost volume and all 3D convolutions are reduced with the same factor. As the number of channels decreasing, our models with group-wise correlation volume (\textit{Gwc-Cat}) perform much better than the models with only concatenation volume (\textit{Cat}). The performance gain enlarges as more channels reduced. The reason for this is that the group-wise correlation provides good matching cost representations for the 3D aggregation network, while the aggregation network with only concatenation volume as inputs needs to learn the matching similarity function from scratch, which usually requires more parameters and computational cost. As a result, the proposed group-wise correlation could be a valuable method to be implemented in real-time stereo networks where the computational costs are limited.

\subsection{KITTI 2012 and KITTI 2015}
For KITTI stereo 2015~\cite{menze2015object}, we split the training set into 180 training image pairs and 20 validation image pairs. Since the results on the validation set are not stable, we fine-tune the pretrained model for 3 times and choose the model with the best validation performance. From Table \ref{table:ablation-kitti}, the performance of both \textit{Gwc40-Cat24} and \textit{Gwc40} is better than the models without group-wise correlation (\textit{Cat64}, \textit{Cat64-original-hg}). We submit the \textit{Gwc40} model (without concatenation volume) with the lowest validation error to the evaluation server, and the results on the test set are shown in Table \ref{table:result-kitti2015}. Our model surpasses the PSMNet~\cite{chang2018pyramid} by 0.21\% and SegStereo~\cite{yang2018segstereo} by 0.14\% on D1-all.

For KITTI 2012~\cite{Geiger2012CVPR}, we split the training set into 180 training images and 14 validation image pairs. The results on the validation set are shown in Table \ref{table:ablation-kitti}. We submit the best \textit{Gwc40-Cat24} model on the validation set to the evaluation server. The evaluation results on the test set are shown in Table \ref{table:result-kitti2012}. Our method surpasses PSMNet~\cite{chang2018pyramid} by 0.19\% on 3-pixel-error and 0.1px on mean disparity error.

%%%%%%%%%%%%%%%%%%%%%%%%%%%%%%%%%%%%%%%%%%%%%%%%%%%%%%%%%%%%%
%%%%%%%%%%%%%%%%%%%%%%%%% Conclusion
%%%%%%%%%%%%%%%%%%%%%%%%%%%%%%%%%%%%%%%%%%%%%%%%%%%%%%%%%%%%%
\section{Conclusion}
In this paper, we proposed GwcNet to estimate disparity maps for stereo matching, which incorporates group-wise correlation to build up the cost volumes. The group-wise correlation volumes provide good matching features for the 3D aggregation network, which improves the performance and reduces the parameter requirements of the aggregation network. We showed that when the computational cost is limited, our model achieves larger gain than previous concatenation-volume based stereo networks. We also improved the stacked hourglass networks to further improve the performance and reduce the inference time. Experiments demonstrated the effectiveness of our proposed method on the Scene Flow datasets and the KITTI dataset.

{\small
\bibliographystyle{ieee}
\bibliography{citations}
}

\end{document}